\documentclass[10pt,twocolumn,letterpaper]{article}

\usepackage{iccv}
\usepackage{times}
\usepackage{epsfig}
\usepackage{graphicx}
\usepackage{amsmath}
\usepackage{amssymb}
\usepackage{mathrsfs}
\usepackage{mathtools}
\usepackage{subfigure}
\usepackage{multirow}

\DeclareMathOperator*{\argmin}{arg\,min}

\usepackage[pagebackref=true,breaklinks=true,letterpaper=true,colorlinks,bookmarks=false]{hyperref}

\iccvfinalcopy 

\def\iccvPaperID{1416} 

\ificcvfinal\fi
\begin{document}

\title{Multi-scale recognition with DAG-CNNs}

\author{Songfan Yang\\
College of Electronics and Information Engineering,\\
Sichuan University, China\\
{\tt\small syang@scu.edu.cn}
\and
Deva Ramanan\\
Deptment of Computer Science,\\
University of California, Irvine, USA\\
{\tt\small dramanan@ics.uci.edu}
}

\maketitle

\begin{abstract}
We explore multi-scale convolutional neural nets (CNNs) for image classification. Contemporary approaches extract features from a single output layer. By extracting features from multiple layers, one can simultaneously reason about high, mid, and low-level features during classification. The resulting multi-scale architecture can itself be seen as a feed-forward model that is structured as a directed acyclic graph (DAG-CNNs). 
We use DAG-CNNs to learn a set of multiscale features that can be effectively shared between coarse and fine-grained classification tasks. While fine-tuning such models helps performance, we show that even ``off-the-self'' multiscale features perform quite well. We present extensive analysis and demonstrate state-of-the-art classification performance on three standard scene benchmarks (SUN397, MIT67, and Scene15). 
In terms of the heavily benchmarked MIT67 and Scene15 datasets, our results reduce the lowest previously-reported error by {\bf 23.9\%} and {\bf 9.5\%}, respectively.
\end{abstract}

\section{Introduction}

\begin{figure}[t!]
\centering
\includegraphics[width=\columnwidth]{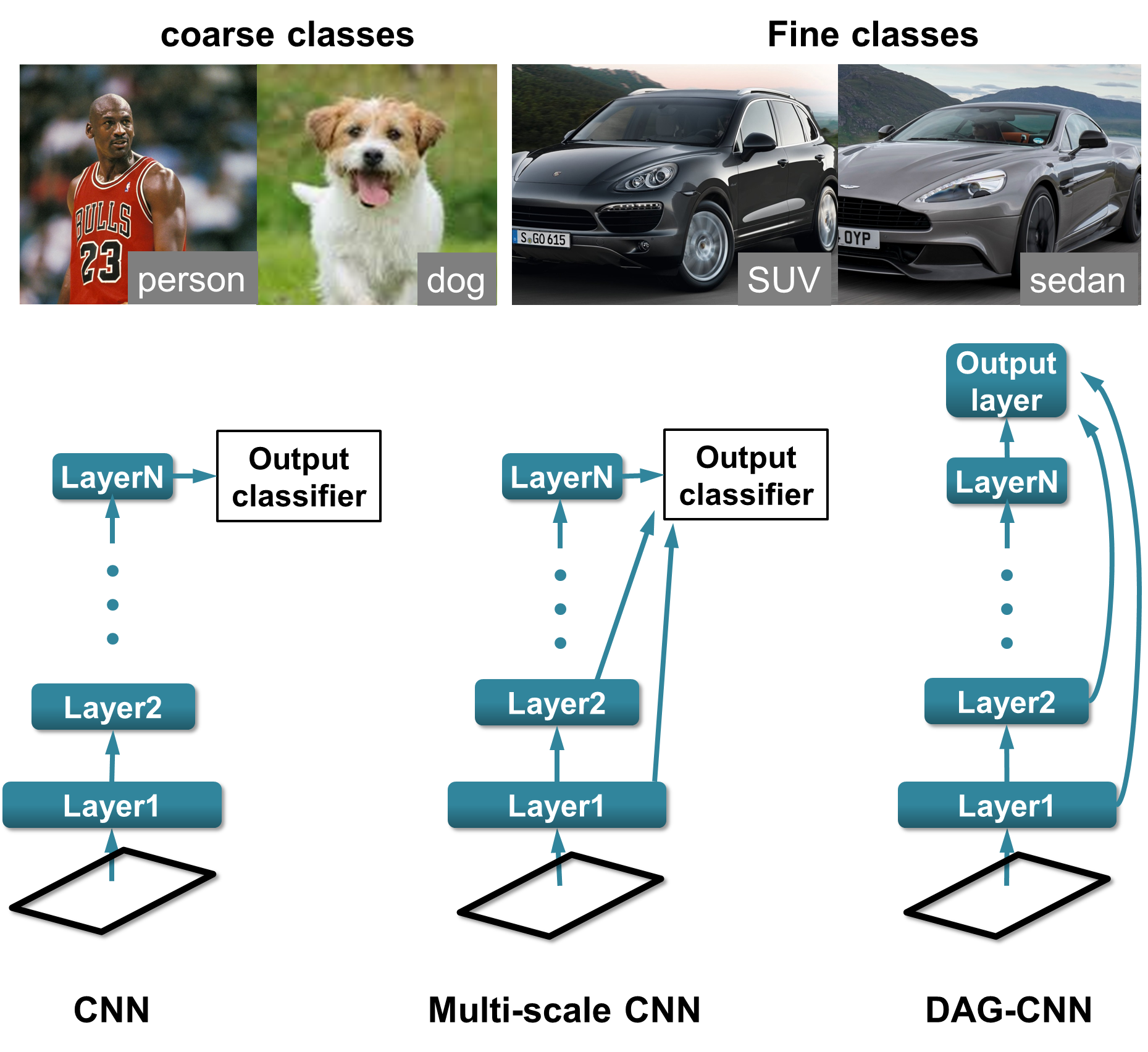}
\caption{Recognition typically require features at multiple scales. Distinguishing a person versus dog requires highly invariant features robust to the deformation of each category. On the other hand, fine-grained recognition likely requires detailed shape cues to distinguish models of cars ({\bf top}). We use these observations to revisit deep convolutional neural net (CNN) architectures. Typical approaches train a classifier using features from a single output layer ({\bf left}). We extract multi-scale features from multiple layers to simultaneously distinguish coarse and fine classes. Such features come ``for free'' since they are already computed during the feed-forward pass ({\bf middle}). Interestingly, the entire multi-scale predictor is still a feed-forward architecture that is no longer chain-structured, but a directed-acyclic graph (DAG) ({\bf right}). We show that DAG-CNNs can be discriminatively trained in an end-to-end fashion, yielding state-of-the-art recognition results across various recognition benchmarks. }
\label{fig:splash}
\end{figure}

Deep convolutional neural nets (CNNs), pioneered by Lecun and collaborators~\cite{lecun1998gradient}, now produce state-of-the-art performance on many visual recognition tasks~\cite{AlexNet, veryDeep, GoogLeNet}. An attractive property is that appear to serve as universal feature extractors, either as ``off-the-shelf'' features or through a small amount of ``fine tuning''. CNNs trained on particular tasks such as large-scale image classification~\cite{ImageNet} {\em transfer} extraordinarily well to other tasks such as object detection~\cite{rcnn}, scene recognition~\cite{zhoulearning}, image retrieval~\cite{Gong14}, etc \cite{cnn_baseline}.

{\bf Hierarchical chain models:}  CNNs are 
hierarchical feed-forward architectures that compute progressively invariant representations of the input image. However, the appropriate level of invariance might be task-dependent. Distinguishing people and dogs requires a representation that is robust to large spatial deformations, since people and dogs can articulate. However, fine-grained categorization of car models (or bird species) requires fine-scale features that capture subtle shape cues. We argue that a universal architecture capable of both tasks must employ some form of multi-scale features for output prediction.

\begin{figure*}
\centering
	\subfigure[mid-level feature is preferred]{\includegraphics[width=.7\textwidth]{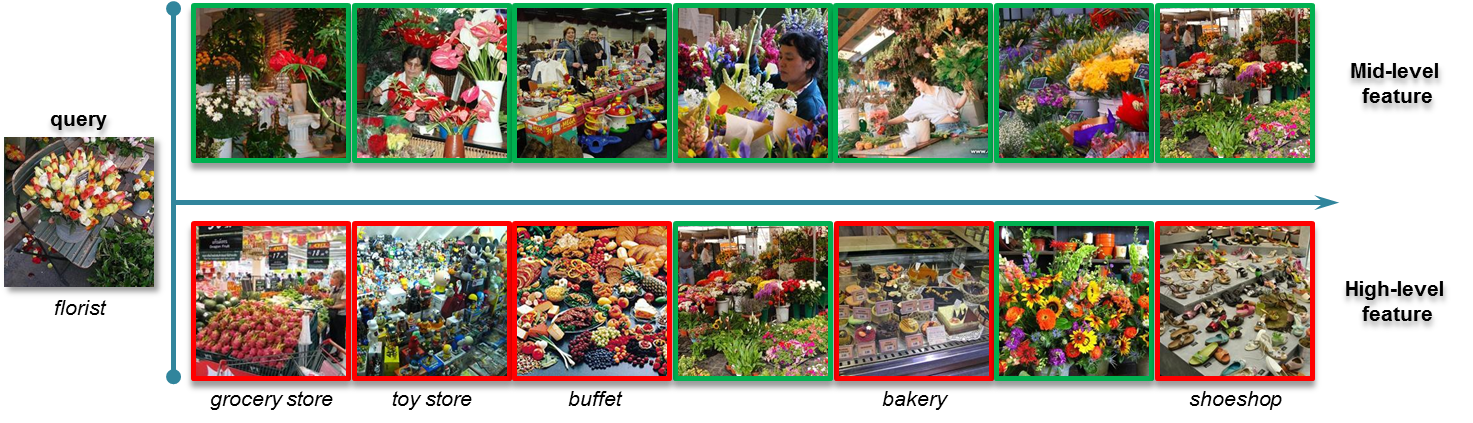}\label{fig:moti_low}}
	\subfigure[high-level feature is preferred]{\includegraphics[width=.7\textwidth]{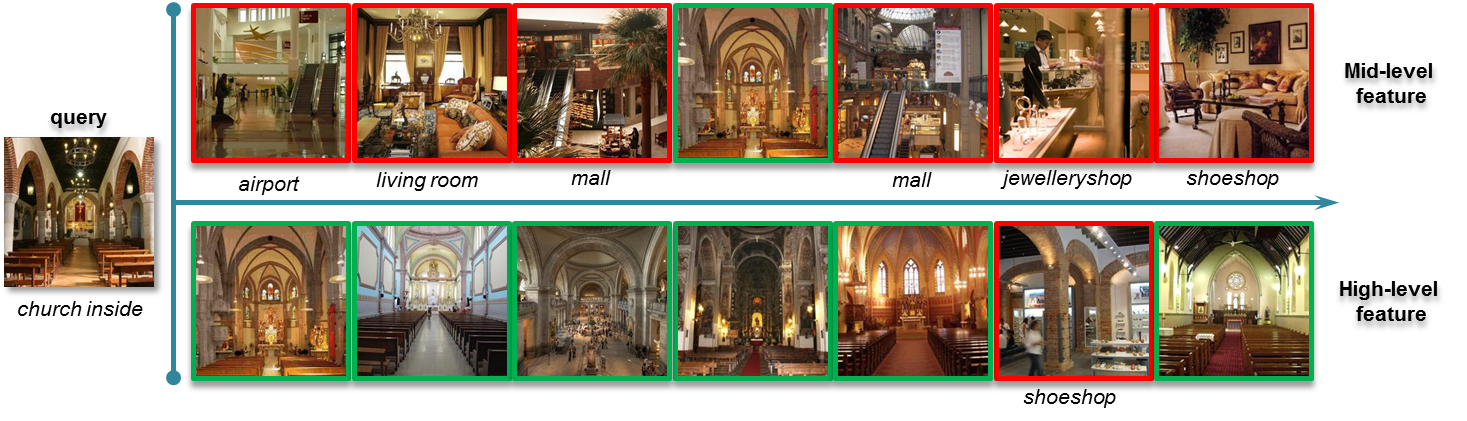}\label{fig:moti_high}}
\caption{Retrieval results using L2 distance for both mid- and high-level features on MIT67~\cite{MIT67}, computed from layer 11 and 20 of the Caffe model. \textit{Green} (\textit{Red}) box means correct (wrong) results, in terms of the scene category label. The correct label for wrong retrievals are provided. The retrieval results are displayed such that the left-most image has the closest distance to the query, and vice versa. Certain query images (or categories) produce better matches with high-level features, while others produce better results with mid-level features. This motivates our multi-scale approach.}
\label{fig:moti}
\end{figure*}

{\bf Multi-scale representations:}  Multi-scale representations are a classic concept in computer vision, dating back to image pyramids~\cite{burt1983laplacian}, scale-space theory ~\cite{lindeberg1993scale}, and multiresolution models~\cite{mallat1999wavelet}. Though somewhat fundamental notions, they have not been tightly integrated with contemporary feed-forward approaches for recognition. We introduce multi-scale CNN architectures that use features at multiple scales for output prediction (Fig.~\ref{fig:splash}). From one perspective, our architectures are quite simple. Typical approaches train a output predictor (e.g., a linear SVM) using features extracted from a single output layer. Instead, one can train an output predictor using features extracted from {\em multiple} layers. Note that these features come ``for free''; they are already computed in a standard feed-forward pass.

{\bf Spatial pooling:} One difficulty with multi-scale approaches is feature dimensionality - the total number of features across all layers can easily reach hundreds of thousands. This makes training even linear models difficult and prone to over-fitting. Instead, we use marginal activations computed from sum (or max) pooling across spatial locations in a given activation layer. From this perspective, our models are similar to those that compute multi-scale features with spatial pooling, including multi-scale templates~\cite{felzenszwalb2008discriminatively}, orderless models\cite{Gong14}, spatial pyramids~\cite{spatial_pyramid}, and bag-of-words~\cite{sivic2003video}. Our approach is most related to \cite{Gong14}, who also use spatially pooled CNN features for scene classification. They do so by pooling together multiple CNN descriptors (re)computed on various-sized patches within an image. Instead, we perform a single CNN encoding of the entire image, extracting multiscale features ``for free''.

{\bf End-to-end training:} Our multi-scale model differs from such past work in another notable aspect. Our entire model is still a feed-forward CNN that is no longer chain-structured, but a directed-acyclic graph (DAG). DAG-structured CNNs can still be discriminatively trained in an end-to-end fashion, allowing us to directly learn multi-scale representations. 
DAG structures are relatively straightforward to implement given the flexibility of many deep learning toolboxes~\cite{vedaldimatconvnet,Caffe}. Our primary contribution is the demonstration that structures can capture multiscale features, which in turn allow for transfer learning between coarse and fine-grained classification tasks.

{\bf DAG Neural Networks:} DAG-structured neural nets were explored earlier in the context of recurrent neural nets \cite{baldi2003principled,graves2009offline}. Recurrent neural nets use feedback to capture dynamic states, and so typically cannot be processed with feed-forward computations. 
More recently, networks have explored the use of ``skip'' connections between layers \cite{raiko-aistats-12,GoogLeNet,sermanet2013pedestrian}, similar to our multi-scale connections. \cite{raiko-aistats-12} show that such connections are useful for a single binary classification task, but we motivate multiscale connections through multitask learning: different visual classification tasks require features at different image scales. 
Finally, the recent state-of-the-art model of \cite{GoogLeNet} make use of skip connections for training, but does not use them at test-time. This means that their final feedforward predictor is not a DAG. Our results suggest that adding multiscale connections at testtime might further improve their performance.

{\bf Overview:} We motivate our multi-scale DAG-CNN model in Sec.~\ref{sec:motivation}, describe the full architecture in Sec.~\ref{sec:approach}, and conclude with numerous benchmark results in Sec.~\ref{sec:exp}. We evaluate multi-scale DAG-structured variants of existing CNN architectures (\eg, Caffe~\cite{Caffe}, Deep19~\cite{veryDeep}) on a variety of scene recognition benchmarks including SUN397~\cite{SUN397}, MIT67~\cite{MIT67}, Scene15~\cite{Scene15}. We observe a consistent improvement regardless of the underlying CNN architecture, producing state-of-the-art results on all 3 datasets.


\section{Motivation\label{sec:motivation}}

In this section, we motivate our multi-scale architecture with a series of empirical analysis. We carry out an analysis on existing CNN architectures, namely Caffe and Deep19. Caffe~\cite{Caffe} is a broadly used CNN toolbox. It includes a pre-trained model ``AlexNet''~\cite{AlexNet} model, learned with millions of images from the ImageNet dataset~\cite{ImageNet}. This model has 6 conv. layers and 2 fully-connected (FC) layers. Deep19~\cite{veryDeep} uses very small $3\times 3$ receptive fields, but an increased number of layers -- 19 layers (16 conv. and 3 FC layers). This model produced state-of-the-art performance in ILSVRC-2014 classification challenge~\cite{ILSVRC14}. We evaluate both ``off-the-shelf'' pre-trained models on the heavily benchmarked MIT Indoor Scene (MIT67) dataset~\cite{MIT67}, using 10-fold cross-validation.

\subsection{Single-scale models} 

{\bf Image retrieval:} Recent work has explored sparse reconstruction techniques for visualizing and analyzing features~\cite{vondrick2013hoggles}. Inspired by such techniques, we use image retrieval to begin our exploration. We attempt to ``reconstruct'' a query image by finding $M=7$ closest images in terms of L2-distance, when computed with mean-pooled layer-specific activations. Results are shown for two query images and two Caffe layers in Fig.~\ref{fig:moti}. The {\tt florist} query image tends to produces better matches when using mid-level features that appear to capture \textit{objects} and \textit{parts}. On the other hand, the {\tt church-inside} query image tends to produce better matches when using high-level features that appear to capture more global \textit{scene} statistics.



\begin{figure}[t!]
\centering
	\includegraphics[width=\columnwidth]{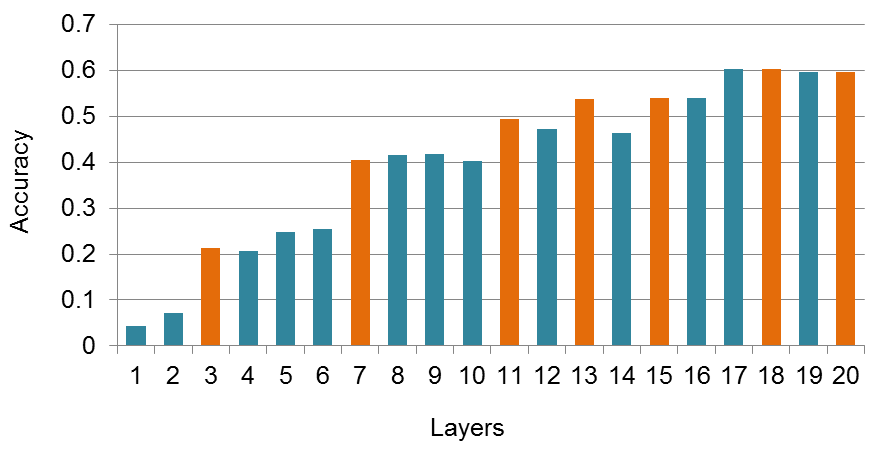}

\caption{The classification accuracy on MIT67~\cite{MIT67} using activations from each layer. We use a solid color fill representing the output of a ReLU layer, where there are 7 in total for the Caffe model. We tend to see a performance jump at each successive ReLU layer, particularly earlier on in the model.}
\label{fig:layer_MIT67}
\end{figure}

{\bf Single-scale classification:} Following past work \cite{cnn_baseline}, we train a linear SVM classifier using features extracted from a particular layer. We specifically train $K=67$ 1-vs-all linear classifiers.
We plot the performance of single-layer classifiers in Fig.~\ref{fig:layer_MIT67}. The detailed parameter options for both Caffe model are described later in Sec.~\ref{sec:exp}. As past work has pointed out, we see a general increase in performance as we use higher-level (more invariant) features. We do see a slight improvement at each nonlinear activation (ReLU) layer. This makes sense as this layer introduces a nonlinear rectification operation $\max(0,x)$, while other layers (such an convolutional or sum-pooling) are linear operations that can be learned by a linear predictor.

\begin{figure*}[ht!]
\centering
	\includegraphics[width=1\textwidth]{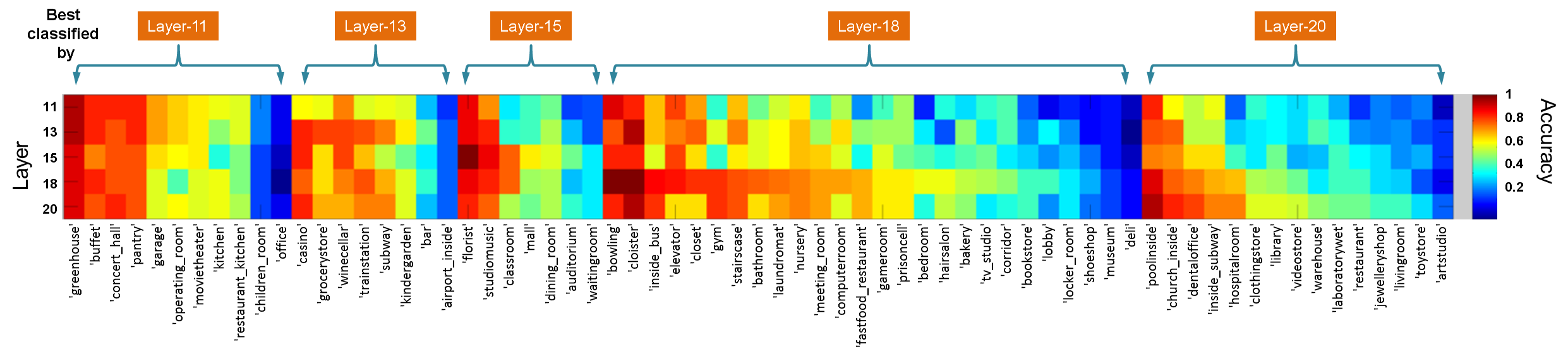}
\caption{The ``per-class'' performance using features extracted from particular layers. We group classes with identical best-performing layers. The last layer ({\bf 20}) is optimal for only 15 classes, while the second-to-last layer ({\bf 18}) proves most discriminative for 26 classes. The third-most effective layer ({\bf 11}) captures significantly lower-level level features. These results validate our underlying hypothesis; different classes require different amounts of invariance. This suggests that a feature extractor shared across such classes will be more effective when multi-scale.}
\label{fig:level_perf}
\end{figure*}


{\bf Scale-varying classification:} The above experiment required training $K \times N$ 1-vs-all classifiers, where $K$ is the number of classes and $N$ is the number of layers. We can treat each of the $KN$ classifiers as binary predictors, and score each with the number of correct detections of its target class. We plot these scores as a matrix in Fig.~\ref{fig:level_perf}. We tend to see groups of classes that operate best with features computed from particular high-level or mid-level layers. 
Most categories tend to do well with high-level features, but a significant fraction (over a third) do better with mid-level features.

\begin{figure}[t!]
\centering
	{\includegraphics[width=.6\columnwidth]{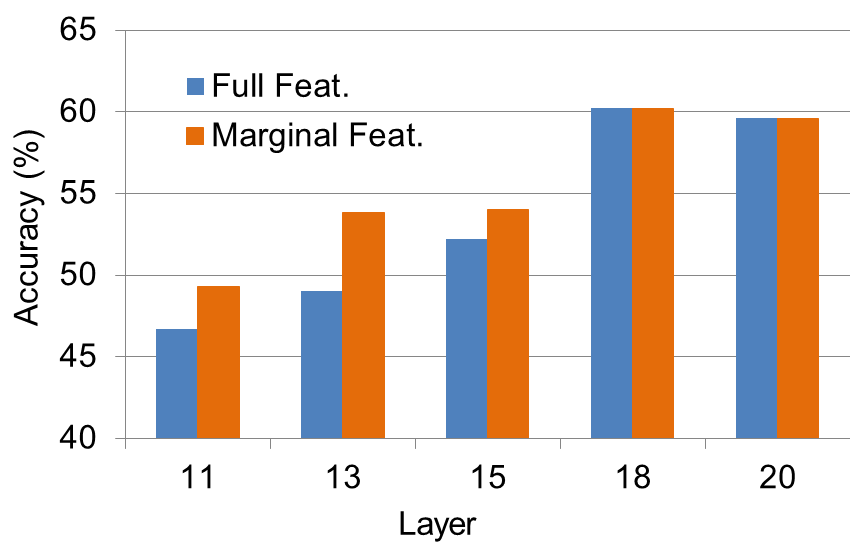}}
\caption{Marginal features (computed by spatially-pooling across activations from a layer) do as well or better than their full-dimensional counterpart. }
\label{fig:full_marg}
\end{figure}

{\bf Spatial pooling:} In the next section, we will explore multi-scale features. One practical hurdle to including all features from all layers is the massive increase in dimensionality. Here, we explore strategies for reducing dimensionality through pooled features. We consider various pooling strategies (sum, average, and max), pooling neighborhoods, and normalization post-processing (with and without L2 normalization). We saw good results with average pooling over all spatial locations, followed by L2 normalization. Specifically, assume a particular layer is of size $H \times W \times F$, where $H$ is the height, $W$ is the width, and $F$ is the number of filter channels. We compute a $1 \times 1 \times F$ feature by averaging across spatial dimensions. We then normalize this feature to have unit norm. 
We compare this encoding versus the unpooled full-dimensional feature (also normalized) in Fig.~\ref{fig:full_marg}. Pooled features always do better, implying dimensionality reduction actually helps performance. We verified that this phenomena was due to over-fitting; the full features always performed better on training data, but performed worse on test data. This suggests that with additional training data, less-aggressive pooling (that preserves some spatial information) may perform better. 

\subsection{Multi-scale models} 

\begin{figure}[t!]
\centering
	\subfigure[Multi-scale]{\includegraphics[width=.49\columnwidth]{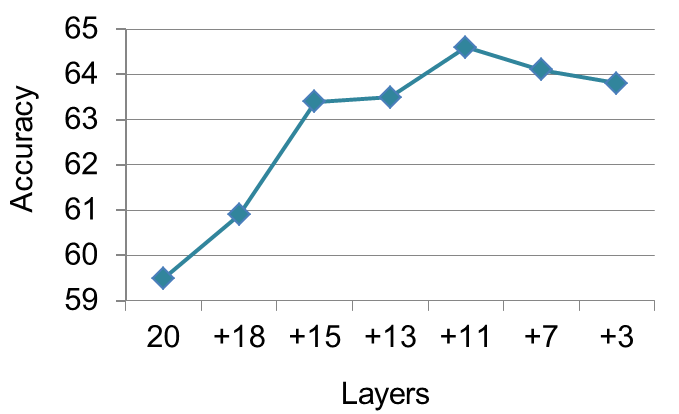}\label{fig:add_back_caffe}}
	\subfigure[Forward Selection]{\includegraphics[width=.49\columnwidth]{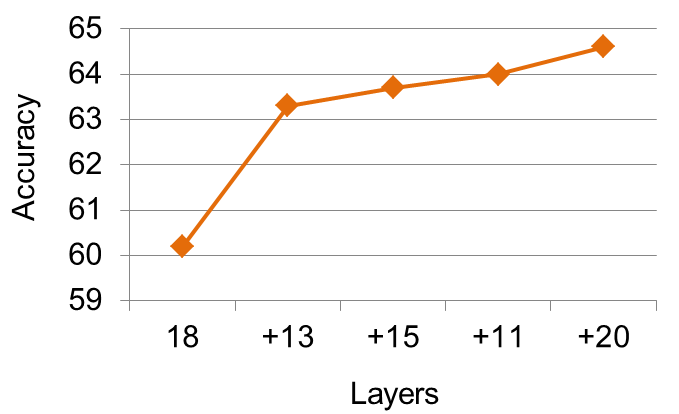}\label{fig:forward_select_caffe}}
\caption{ (a) The performance of a multi-scale classifier as we add more layer-specific features. We start with the last ReLU layer, and iteratively add the previous ReLU layer to the feature set of the classifier.
The ``+'' sign means the recent-most added layer. Adding additional layers help, but performance saturates and even slightly decreases when adding lower-layer features. This suggests it may be helpful to search for the ``optimal'' combination of layers. (b) The performance trend when using forward selection to incorporate the ReLU layers. Note that layers are {\em not} selected in high-to-low order. Specifically, it begin with the second-to-last ReLU layer, and skip one or more previous layers when adding the next layer. This suggests that layers encode some redundant or correlated information. Overall, we see a significant 6\% improvement.}
\label{fig:add_back}
\end{figure}

{\bf Multiscale classification:} We now explore multi-scale predictors that process pooled features extracted from multiple layers. As before, we analyze ``off-the-shelf'' pre-trained models. We evaluate performance as we iteratively add more layers. Fig.~\ref{fig:layer_MIT67} suggests that the last ReLU layer is a good starting point due to its strong single-scale performance. Fig~\ref{fig:add_back_caffe} plots performance as we add previous layers to the classifier feature set. Performance increases as we add intermediate layers, while lower layers prove less helpful (and may even hurt performance, likely do to over-fitting). Our observations suggest that high and mid-level features (\ie, \textit{parts} and \textit{objects}) are more useful than low-features based on \textit{edges} or \textit{textures} in scene classification.

{\bf Multi-scale selection:} The previous results show that adding all layers may actually hurt performance. We verified that this was an over-fitting phenomena; additional layers always improved training performance, but could decrease test performance due to over-fitting. This appears especially true for multi-scale analysis, where nearby layers may encoded redundant or correlated information (that is susceptible to over-fitting). Ideally, we would like to search for the ``optimal'' combination of ReLU layers that maximize performance on validation data. Since there exists an exponential number of combinations ($2^N$ for $N$ ReLU layers), we find an approximate solution with a greedy forward-selection strategy. We greedily select the next-best layer (among all remaining layers) to add, until we observe no further performance improvement. As seen in Fig.~\ref{fig:forward_select_caffe}, the optimal results of this greedy approach rejects the low-level features. This is congruent with the previous results in Fig.~\ref{fig:add_back_caffe}. 

Our analysis strongly suggest the importance (and ease) of incorporating multi-scale features for classification tasks. For our subsequent experiments, we use scales selected by the forward selection algorithm on MIT67 data (shown in Fig.~\ref{fig:forward_select_caffe}). Note that we use them for all our experimental benchmarks, demonstrating a degree of cross-dataset generalization in our approach. 

\section{Approach\label{sec:approach}} 

\begin{figure*}[t!]
\centering
	\includegraphics[width=.9\textwidth]{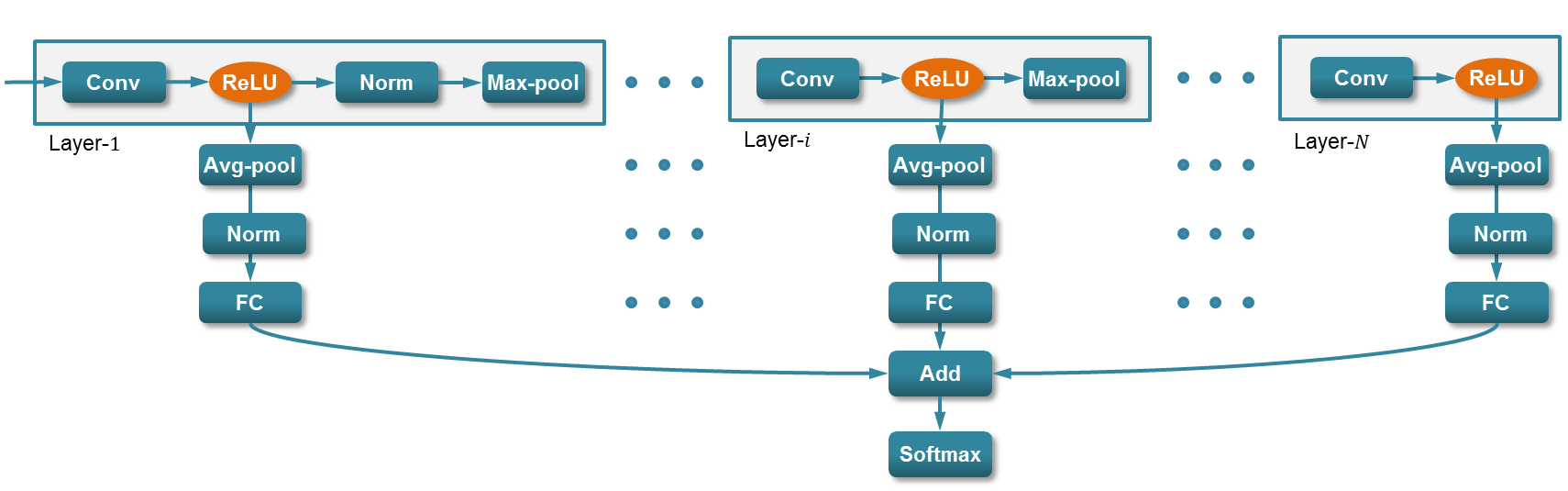}
\caption{Our multi-scale DAG-CNN architecture is constructed by adding multi-scale output connections to an underlying {\em chain backbone} from the original CNN. Specifically, for each scale, we spatially (average) pool activations, normalize them to have unit-norm, compute an inner product with a fully-connected (FC) layer with $K$ outputs, and add the scores across all layers to predictions for $K$ output classes (that are finally soft-maxed together).}
\label{fig:model}
\end{figure*}

In this section, we show that the multi-scale model examined in Fig.~\ref{fig:forward_select_caffe} can be written as a DAG-structured, feed-forward CNN. Importantly, this allows for end-to-end gradient-based learning. To do so, we use standard calculus constructions -- specifically the chain rule and partial derivatives -- to generalize back-propagation to layers that have multiple ``parents'' or inputs. Though such DAG structures have been previously introduced by prior work, we have not seen derivations for the corresponding gradient computations. We include them here for completeness, pointing out several opportunities for speedups given our particular structure.

{\bf Model:} The run-time behavior of our multi-scale predictor from the previous section is equivalent to feed-forward processing of the DAG-structured architecture in Fig.~\ref{fig:forward_select_caffe}. Note that we have swapped out a margin-based hinge-loss (corresponding to a SVM) with a softmax function, as the latter is more amenable to training with current toolboxes. Specifically, typical CNNs are grouped into collections of four layers, \ie, Conv., ReLU, contrast normalization (Norm), pooling layers (with the Norm and pooling layers being optional). The final layer is usually a $K$-way softmax function that predicts one of $K$ outputs. We visualize these layers as a chain-structured ``backbone'' in Fig.~\ref{fig:model}. Our DAG-CNN simply links each ReLU layer to an average-pooling layer, followed by a L2 normalization layer, which feeds to a fully-connected (FC) layer that produces $K$ outputs (represented formally as a $1 \times 1 \times K$ matrix). These outputs are element-wise added together across all layers, and the resulting $K$ numbers are fed into the final softmax function. The weights of the FC layers are equivalent to the weights of the final multi-scale $K$-way predictor (which is a softmax predictor for a softmax loss output, and a SVM for a hinge-loss output). Note that all the required operations are standard modules except for the \textit{Add}.

{\bf Training:} Let $\textbf{w}_1,...\textbf{w}_K$ be the CNN model parameters at $1,..,K$-th layer, training data be ($\textbf{x}^{(i)},\textbf{y}^{(i)}$), where $\textbf{x}^{(i)}$ is the $i$-th input image and $\textbf{y}^{(i)}$ is the indicator vector of the class of $\textbf{x}^{(i)}$. Then we intend to solve the following optimization problem

\begin{align}
\argmin_{\textbf{w}_1,...\textbf{w}_K} \frac{1}{n}\sum_{i=1}^{n} \mathcal{L}(f(\textbf{x}^{(i)};\textbf{w}_1,...,\textbf{w}_K),\textbf{y}^{(i)})
\end{align}

As is now commonplace, we make use of stochastic gradient descent to minimize the objective function. For a traditional \textit{chain} model, the partial derivative of the output with respect to any one weight can be recursively computed by the chain rule, as described in the back-prop algorithm.

\begin{figure}[t!]
\centering
	\includegraphics[width=\columnwidth]{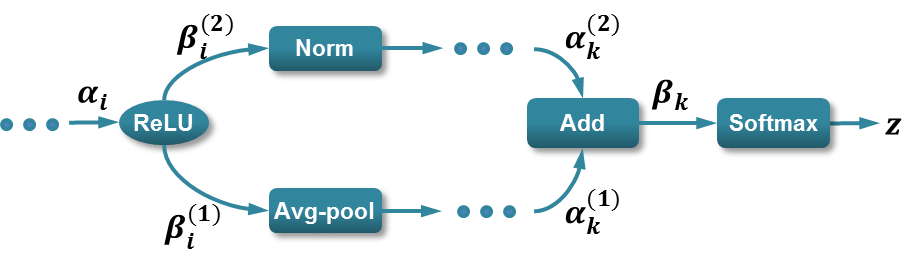}
\caption{Visualization of the parameter setup at $i$-th ReLU.}
\label{fig:backprop_eq}
\end{figure}

{\bf Multi-output layers (ReLU):} Our DAG-model is structurally different at the ReLU layers (since they have multiple outputs) and the \textit{Add} layer (since it has multiple inputs). We can still compute partial derivatives by recursively applying the chain rule, but care needs to be taken at these points. Let us consider the $i$-th ReLU layer in Fig.~\ref{fig:backprop_eq}. Let $\alpha_i$ be its input, $\beta_i^{(j)}$ be the output for its $j$-th output branch (its $j^{th}$ child in the DAG), and let $z$ is the final output of the softmax layer. The gradient of $z$ with respect to the input of the $i$-th ReLU layer can be computed as

\begin{align}
\frac{\partial z}{\partial \alpha_i}=\sum_{j=1}^{C}\frac{\partial z}{\partial \beta_i^{(j)}}\frac{\partial \beta_i^{(j)}}{\partial \alpha_i} \quad \text {(in general)} \label{eq:backprop1}
\end{align}

\noindent where $C=2$ for the example in Fig.~\ref{fig:backprop_eq}. One can recover standard back-propagation equations from the above by setting $C=1$: a single back-prop signal $\frac{\partial z}{\partial \beta_i^{(1)}}$  arrives at ReLU unit $i$, is multiplied by the local gradient $\frac{\partial \beta_i^{(1)}}{\partial \alpha_i^{(1)}}$, and is passed on down to the next layer below. In our DAG, {\em multiple} back-prop signals arrive $\frac{\partial z}{\partial \beta_i^{(j)}}$ from each branch $j$, each is multiplied by an branch-specific gradient $\frac{\partial \beta_i^{(j)}}{\partial \alpha_i}$, and their total sum is passed on down to the next layer.


{\bf Multi-input layers (Add):} Let $\beta_k=g(\alpha^{(1)}_k,\cdots,\alpha^{(N)}_k)$ represents the output of a layer with multiple inputs. We can compute the gradient along the layer by applying the chain rule as follows:
\begin{align}
\frac{\partial z}{\partial \alpha_i}&=\frac{\partial z}{\partial \beta_k}\frac{\partial \beta_k}{\partial \alpha_i} \nonumber \\
&=\frac{\partial z}{\partial \beta_k}\sum_{j=1}^{C}\frac{\partial \beta_k}{\partial \alpha_k^{(j)}}\frac{\partial \alpha_k^{(j)}}{\partial \alpha_i} \quad \text{(in general)} \label{eq:backprop2}
\end{align} 
One can similarly arrive at the standard back-propagation by setting $C=1$.

{\bf Special case (ReLU):} Our particular DAG architecture can further simplify the above equations. Firstly, it may be common for multiple-output layers to duplicate the same output for each child branch. This is true of our ReLU units; they pass the same values to the next layer in the chain and the current-layer pooling operation. This means the output-specific gradients are identical for those outputs $\forall j, \frac{\partial \beta_i^{(j)}}{\partial \alpha_i} =  \frac{\partial \beta_i^{(1)}}{\partial \alpha_i}$, which simplifies \eqref{eq:backprop1} to
\begin{align}
\frac{\partial z}{\partial \alpha_i} = \frac{\partial \beta_i^{(1)}}{\partial \alpha_i} \sum_{j=1}^{C}\frac{\partial z}{\partial \beta_i^{(j)}} \quad \text{(for duplicate outputs)}
\end{align}
This allows us to add together multiple back-prop signals before scaling them by the local gradient, reducing 
the number of multiplications by $C$. We make use of this speed up to train our ReLU layers
. 

{\bf Special case(Add):} Similarly, our multi-input {\tt Add} layer reuses the same partial gradient for each input
$\forall j, \frac{\partial \beta_k}{\partial \alpha_k^{(j)}} = \frac{\partial \beta_k}{\partial \alpha_k^{(1)}}$ which simplifies even further in our case to $1$. The resulting back-prop equations that simplify \eqref{eq:backprop2} are given by
\begin{align}
\frac{\partial z}{\partial \alpha_i} = \frac{\partial z}{\partial \beta_k} \frac{\partial \beta_k}{\partial \alpha_k^{(1)}}\sum_{j=1}^{C} \frac{\partial \alpha_k^{(j)}}{\partial \alpha_i}  \quad \text{(for duplicate gradients)}
\end{align}
\noindent implying that one can similarly save $C$ multiplications. The above equations have an intuitive implementation; the standard chain-structured back-propagation signal is simply replicated along each of the parents of the {\tt Add} layer.

{\bf Implementation:} We use the excellent MatConNet codebase to implement our modifications \cite{vedaldimatconvnet}. We implemented a custom {\tt Add} layer and a custom DAG data-structure to denote layer connectivity. Training and testing is essentially as fast as the chain model.

\begin{figure}[htbp]
\centering
\includegraphics[width=.9\columnwidth]{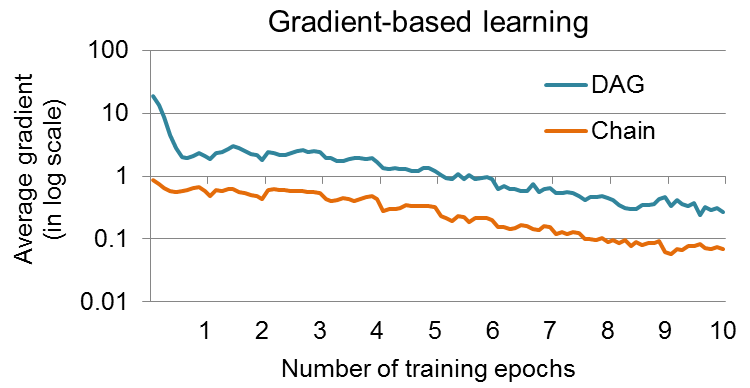}
\caption{The average gradient from Layer-1 (Conv.) during training, plotted in log-scale. Gradients from the DAG are consistently $10\times$ larger, implying that they receive a stronger supervised signal from the target label during gradient-based learning.}
\label{fig:grad}
\end{figure}

{\bf Vanishing gradients:} We point out an interesting property of our multiscale models that make them easier to train. Vanishing gradients~\cite{bengio1994learning} refers to the phenomena that gradient magnitudes decrease as they are propogated through layers, implying that lower-layers can be difficult to learn because they receive too small a learning signal. In our DAG-CNNs, lower layers are {\em directly} connected to the output layer through multi-scale connections, ensuring they receive a strong gradient signal during learning. Fig.~\ref{fig:grad} experimentally verifies this claim.

\section{Experimental Results\label{sec:exp}}
We explore DAG-structured variants of two popular deep models, Caffe~\cite{Caffe} and Deep19~\cite{veryDeep}. We refer to these models as Caffe-DAG and Deep19-DAG. We evaluate these models on three benchmark scene datasets: SUN397~\cite{SUN397}, MIT67~\cite{MIT67}, and Scene15~\cite{Scene15}. In absolute terms, we achieve the best performance ever reported on all three benchmarks, sometimes by a significant margin.


\begin{table*}[t!]
\centering
\resizebox{.9\textwidth}{!}{
\begin{tabular}{ccc}
SUN397 & MIT67 & Scene15\\
\begin{tabular}[t]{|l|c|}
\hline
Approach & Accuracy(\%) \\
\hline
Deep19-DAG & \textbf{56.2} \\
Deep19~\cite{veryDeep} & 51.9 \\
Caffe-DAG & 46.6	\\
Caffe~\cite{Caffe} & 43.5 \\ \hline
Places~\cite{zhoulearning}	& 54.3	\\
MOP-CNN~\cite{Gong14} & 52.0 \\
FV~\cite{FV} & 47.2 \\
DeCaf~\cite{DeCaf} & 40.9	\\
Baseline-overfeat~\cite{SUN_ijcv}	&40.3 \\
Baseline~\cite{SUN397} & 38.0 \\
\hline
\end{tabular}
&
\begin{tabular}[t]{|l|c|}
\hline
Approach & Accuracy(\%) \\
\hline
Deep19-DAG & \textbf{77.5} \\
Deep19~\cite{veryDeep} & 70.8 \\
Caffe-DAG & 64.6	\\
Caffe~\cite{Caffe} & 59.5 \\ \hline
MOP-CNN~\cite{Gong14} & 68.9 \\
Places~\cite{zhoulearning}	& 68.2	\\
Mid-level~\cite{mid_level} & 64.0	\\
FV+BoP~\cite{FV_BoP} & 63.2 \\
Disc. Patch~\cite{disc_patch} & 49.4 \\
SPM~\cite{spatial_pyramid} & 34.4	\\
\hline
\end{tabular}
&
\begin{tabular}[t]{|l|c|}
\hline
Approach & Accuracy(\%) \\
\hline
Deep19-DAG & \textbf{92.9} \\
Deep19~\cite{veryDeep} & 90.8 \\
Caffe-DAG & 89.7	\\
Caffe~\cite{Caffe} & 86.8 \\ \hline
Place~\cite{zhoulearning} & 91.6 \\
CENTRIST~\cite{Wu_pami11} & 84.8	\\
Hybrid~\cite{Bosch_pami08}	& 83.7	\\
Spatial pyramid~\cite{spatial_pyramid} & 81.4 \\
Object bank~\cite{Li_nips10_objectbank}	& 80.9	\\
Reconfigurable model~\cite{Parizi_cvpr12_reconf} & 78.6	\\
Spatial Envelop~\cite{Oliva_ijcv01_envelop} & 74.1 \\
Baseline~\cite{Scene15} & 65.2 \\
\hline
\end{tabular}
\end{tabular}
}
\caption{Classification results on SUN397, MIT67, and Scene15 datasets. Please see text for an additional discussion.}
\label{table:all}
\end{table*}

{\bf Feature dimensionality:} Most existing methods that use CNNs as feature extractors work with the last layer (or the last fully connected layer), yielding a feature vector of size 4096. 
Forward feature selection on Caffe-DAG selects layers $(11, 13, 15, 18, 20)$, making the final multiscale feature 9216-dimensional. Deep19-DAG selects layers$( 26, 28, 31, 33, 30)$, for a final size of 6144. We perform feature selection by cross-validating on MIT67, and use the same multiscale structure for all other datasets. Dataset-dependant feature selection may further improve performance. Our final multiscale DAG features are {\em only 2X larger} than their single-scale counter part, making them practically easy to use and store.

{\bf Training:}  We follow the standard image pre-processing steps of fixing the input image size to $224\times 224$ by scaling and cropping, and subtracting out the mean RGB value (computed on ImageNet). We initialize filters and biases to their pre-trained values (tuned on ImageNet) and initialize multi-scale fully-connected (FC) weights to small normally-distributed numbers. We perform 10 epochs of learning.

{\bf Baselines:} We compare our DAG models to published results, including two additional baselines. We evaluate the best single-scale ``off-the-shelf'' model, using both Caffe and Deep19. We pass L2-normalized single-scale features to Liblinear~\cite{liblinear} to train $K$-way one-vs-all classifiers with default settings. Finally, Sec.~\ref{sec:diag} concludes with a detailed diagnostic analysis comparing off-the-shelf and fine-tuned versions of chain and DAG structures.


\begin{figure*}[t!]
\centering
	\includegraphics[width=.9\textwidth]{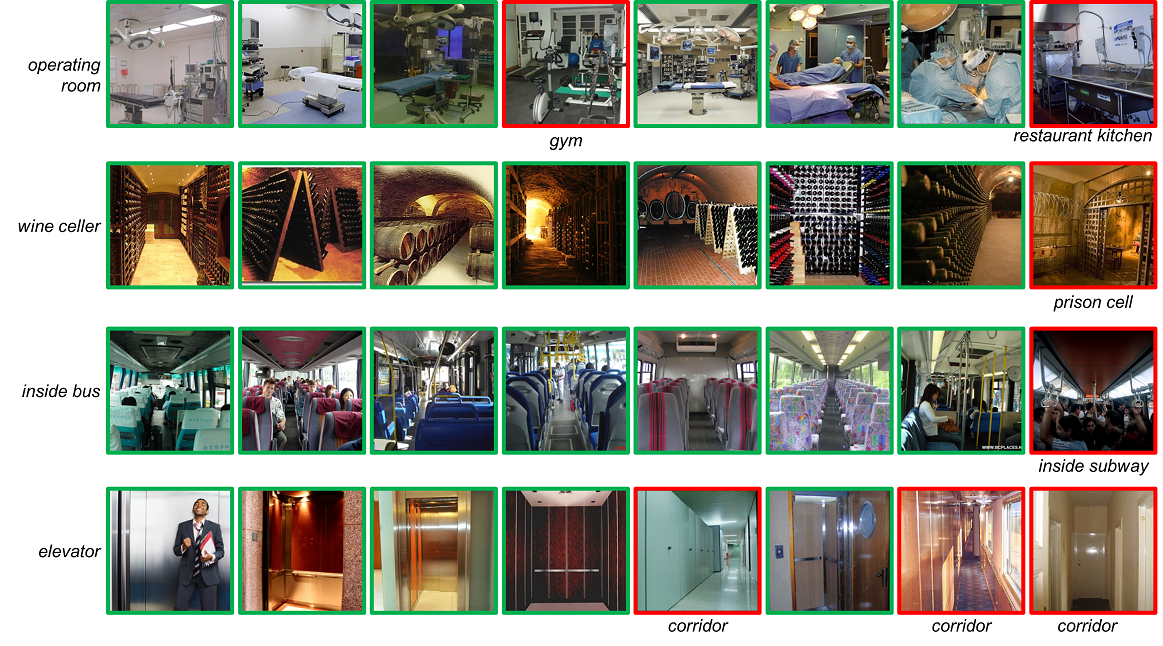}
\caption{Deep19-DAG results on MIT67. The category label is shown on the left and the label for false-positives (in {\bf red}) are also provided. We use the multi-scale analysis of Fig.~\ref{fig:level_perf} to compare categories that perform better with mid-level features ({\bf top 2 rows}) versus high-level features ({\bf bottom 2 rows}). Mid-level features appear to emphasize \textit{objects} such as operating equipment for {\tt operating room} scenes and circular grid for {\tt  wine celler}, while high-level features appear to focus on global \textit{spatial statistics} for {\tt inside bus} and {\tt elevator} scenes. }
\label{fig:more_eg}
\end{figure*}

{\bf SUN397:} We tabulate results for all our benchmark datasets in Table~\ref{table:all}, and discuss each in turn. SUN397~\cite{SUN397} is a large scene recognition dataset with 100K images spanning 397 categories, provided with standard train-test splits. Our DAG models outperform their single-scale counterparts. In particular, Deep19-DAG achieves the highest $56.2\%$ accuracy. Our results are particularly impressive given that the next-best method of~\cite{zhoulearning} (with a score of $54.3$) makes use of a ImageNet-trained CNN and a custom-trained CNN using a new 7-million image dataset with 400 scene categories.


{\bf MIT67:} MIT67 consists of 15K images spanning 67 indoor scene classes~\cite{MIT67}, provided with standard train/test splits. Indoor scenes are interesting for our analysis because some scenes are well characterized by high-level spatial geometry (\eg~{\tt church} and {\tt cloister}), while others are better described by mid-level objects (\eg~{\tt wine celler} and {\tt operating room}) in various spatial configurations. We show qualitative results in Fig.~\ref{fig:more_eg}. Deep19-DAG produces a classification accuracy of $77.5\%$, reducing the best-previously reported error~\cite{Gong14} by {\bf 23.9\%}. Interestingly~\cite{Gong14} also uses multi-scale CNN features, but do so by first extracting various-sized patches from an image, rescaling each to canonical size. Single-scale CNN features extracted from these patches are then vector-quantized into a large-vocabulary codebook, followed by a projection step to reduce dimensionality. Our multi-scale representation, while similar in spirit, is an end-to-end trainable model that is computed ``for free'' from a single (DAG) CNN.



{\bf Scene15:} The Scene15~\cite{Scene15} includes both indoor scene (\eg, store and kitchen) and outdoor scene (\eg, mountain and street). It is a relatively small dataset by contemporary standards (2985 test images), but we include here for completeness. Performance is consistent with the results above. Our multi-scale DAG model, specifically Deep19-DAG, outperforms all prior work. For reference, the next-best method of~\cite{zhoulearning} uses a new custom 7-million image scene dataset for training.



\subsection{Diagnostics \label{sec:diag}}
In this section, we analyze ``off-the-shelf'' (OTS) and ``fine-tuned'' (FT) versions of both single-scale chain and multi-scale DAG models. We focus on the Caffe model, as it is faster and easier for diagnostic analysis. 

{\bf Chain:} Chain-OTS uses single-scale features extracted from CNNs pre-trained on ImageNet. These are the baseline Caffe results presented in the previous subsections. Chain-FT trains a model on the target dataset, using the pre-trained model as an initialization. This can be done with standard software packages~\cite{vedaldimatconvnet}. To ensure consistency of analysis, in both cases features are passed to a K-way multi-class SVM to learn the final predictor. 

{\bf DAG:} DAG-OTS is obtained by fixing all internal filters and biases to their pre-trained values, and only learning the multiscale fully-connected (FC) weights. Because this final stage learning is a convex problem, this can be done by simply passing off-the-shelf multi-scale features to a convex linear classification package (e.g., SVM). We compare this model to a fine-tuned version that is trained end-to-end, making use of the modified backprop equation from Sec.~\ref{sec:approach}. 

\begin{figure}[t]
\centering
	\subfigure[SUN397]{\includegraphics[width=.32\columnwidth]{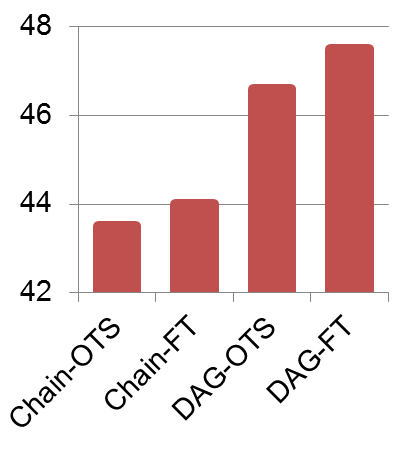}\label{fig:comp_sun}}
	\subfigure[MIT67]{\includegraphics[width=.32\columnwidth]{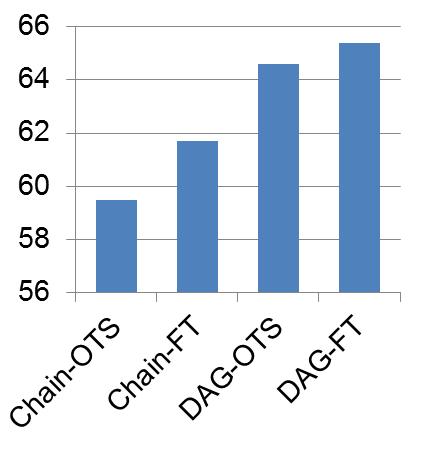}\label{fig:comp_mit}}
	\subfigure[Scene15]{\includegraphics[width=.32\columnwidth]{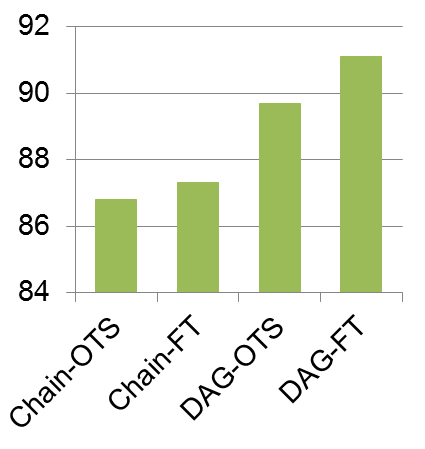}\label{fig:comp_scene}}
	
\caption{Off-the-shelf vs. Fine-tuning models on both Chain and DAG model for Caffe backbone. Please see the text for a discussion.}
\label{fig:comp_otf}
\end{figure}

{\bf Comparison:} Fig.~\ref{fig:comp_otf} compares off-the-shelf and fine-tune variants of chain and DAG models. We see two dominant trends. First, as perhaps expected, fine-tuned (FT) models consistently outperform their off-the-shelf (OTS) countparts. Even more striking is the large improvement from chain to DAG models, indicating the power of multi-scale feature encodings.

{\bf DAG-OTS:} Perhaps most impressive is the strong performance of DAG-OTS. From a theoretical perspective, this validates our underyling hypothesis that multi-scale features allow for better transfer between recognition tasks -- in this case, ImageNet and scene classification. An interesting question is whether multi-scale features, when trained with gradient-based DAG-learning on ImageNet, will allow for even more transfer. We are currently exploring this. However even with current CNN architectures, our results suggest that {\em any system making use of off-the-shelf CNN features should explore multi-scale variants as a ``cheap" baseline.}  Compared to their single-scale counterpart, multiscale features require no additional time to compute, are only a factor of 2 larger to store, and consistently provide a noticeable improvement.

{\bf Conclusion:} We have introduced multi-scale CNNs for image classification. Such models encode scale-specific features that can be effectively shared across both coarse and fine-grained classification tasks. Importantly, such models can be viewed as DAG-structured feedforward predictors, allowing for end-to-end training. While fine-tuning helps performance, we empirically demonstrate that even ``off-the-self'' multiscale features perform quite well. We present extensive analysis and demonstrate state-of-the-art classification performance on three standard scene benchmarks, sometimes improving upon prior art by a significant margin. 
\newpage
{\small
\bibliographystyle{ieee}
\bibliography{mobib}
}

\end{document}